\documentclass{article}

\usepackage{arxiv}

\usepackage[utf8]{inputenc}
\usepackage[T1]{fontenc}    
\usepackage{hyperref}       
\usepackage{url}            
\usepackage{booktabs}      
\usepackage{amsfonts}       
\usepackage{nicefrac}       
\usepackage{microtype}     
\usepackage{lipsum}		
\usepackage{graphicx}
\usepackage{doi}
\usepackage{enumitem}
\usepackage{preamble}
\usepackage{amsmath}
\usepackage{amssymb}
\usepackage{mathtools}
\usepackage{amsthm}
\usepackage{commath}
\usepackage{pifont}
\usepackage[textsize=tiny]{todonotes}
\usepackage{siunitx}
\usepackage{fancyhdr}
\usepackage[numbers]{natbib}

\usepackage{etoolbox,siunitx,booktabs,threeparttable,multirow,array,graphicx}
\usepackage[normalem]{ulem}
\robustify\bfseries
\robustify\uline
\robustify\tnote
\usepackage{multirow}

\usepackage{xcolor}

\author{
\hspace{-0.1cm}Rachael DeVries \\
	University of Copenhagen\\
    Novo Nordisk A/S \\
    rachael.devries@bio.ku.dk
	\And
	Casper Christensen \\
	Independent Researcher\\
\And
	\textbf{Marie Lisandra Zepeda Mendoza} \\
	Novo Nordisk Research Center Oxford Ltd.\\
    \And
	\textbf{Ole Winther} \\
	University of Copenhagen\\ 
    Technical University of Denmark\\
}

\hypersetup{
  colorlinks   = true, 
  urlcolor     = black, 
  linkcolor    = black, 
  citecolor   = black 
}

\renewcommand{\cite}[1]{\citep{#1}}
\newcommand\submittedtext{%
  \footnotesize This work has been submitted to the IEEE for possible publication. Copyright may be transferred without notice, after which this version may no longer be accessible.}

\usepackage{amsmath,amsfonts,bm}

\def\eqref#1{equation~\ref{#1}}

\def\1{\bm{1}}

\def\vh{{\bm{h}}}

\def\vm{{\bm{m}}}

\def\vp{{\bm{p}}}

\def\vx{{\bm{x}}}

\def\mM{{\bm{M}}}

\def\mS{{\bm{S}}}

\def\mX{{\bm{X}}}

\DeclareMathAlphabet{\mathsfit}{\encodingdefault}{\sfdefault}{m}{sl}
\SetMathAlphabet{\mathsfit}{bold}{\encodingdefault}{\sfdefault}{bx}{n}

\newcommand{\R}{\mathbb{R}}

\title{Bi-Axial Transformers: Addressing the Increasing Complexity of EHR Classification}

\begin{document}
\begin{NoHyper}

\maketitle
\pagestyle{fancy}

\fancypagestyle{firstpage}{%
    \fancyhf{} 
    \fancyfoot[C]{\small\submittedtext}
    \renewcommand{\headrulewidth}{0pt}
    \renewcommand{\footrulewidth}{0pt}
}
\thispagestyle{firstpage}

\end{NoHyper}
\begin{abstract}
Electronic Health Records (EHRs), the digital representation of a patient’s medical history, are a valuable resource for epidemiological and clinical research. They are also becoming increasingly complex, with recent trends indicating larger datasets, longer time series, and multi-modal integrations. Transformers, which have rapidly gained popularity due to their success in natural language processing and other domains, are well-suited to address these challenges due to their ability to model long-range dependencies and process data in parallel. But their application to EHR classification remains limited by data representations, which can reduce performance or fail to capture informative missingness. In this paper, we present the Bi-Axial Transformer (BAT), which attends to both the clinical variable and time point axes of EHR data to learn richer data relationships and address the difficulties of data sparsity. BAT achieves state-of-the-art performance on sepsis prediction and is competitive to top methods for mortality classification. In comparison to other transformers, BAT demonstrates increased robustness to data missingness, and learns unique sensor embeddings which can be used in transfer learning. Baseline models, which were previously located across multiple repositories or utilized deprecated libraries, were re-implemented with PyTorch and made available for reproduction and future benchmarking.
\end{abstract}

\fancyhead[C]{\small\submittedtext}



\section{Introduction}
\label{sec:introduction}
Electronic Health Records (EHRs) are being produced in increasing numbers around the world, with many countries incentivizing or requiring their use \cite{Birkhead2015, centers2021national, Slawomirski2023}. This digitalization of healthcare data has given researchers easier access to patient diagnoses, treatments, lab tests, and anthropometric measurements, which can be used to train machine learning models for a variety of clinical applications \cite{nielsen2024, Zang2024, Tang2024}. Classification tasks are popular with EHR data, including predictions of medical codes, hospital readmission, and mortality \cite{med2vec, li2023, AMIRAHMADI2023104430}.

\begin{figure}
    \centering
    \includegraphics[width=88mm]{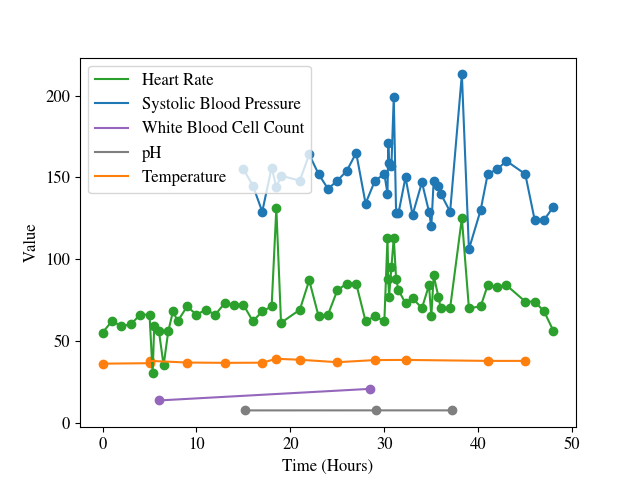}
    \caption{Measurements from five different sensors, generated from a single ICU visit in the Physionet 2012 dataset \cite{goldberger2000physiobank}. Each sensor demonstrates varying levels of sparsity and irregularity.}
    \label{fig:p12_sparsity}
\end{figure}

Many deep learning approaches utilizing EHRs for classification have framed them as time series. Whereas traditional time series data are regular and dense, EHR data are irregular across patients and times, highly sparse, multimodal, and subject to experimental and biological noise (Fig. \ref{fig:p12_sparsity}) \cite{hyndman-athanasopoulos-2018-fpp2, marlin2012, Johnson2016} . A variety of successful models have been developed to address one or more of these characteristics \cite{XIE2022103980}.

Nevertheless, additional challenges are present, accompanying the release of datasets with large numbers of patients, increasingly long histories, and more modalities \cite{DNPR2015, MIMIC-IV, allofus-NIH}. Modeling patients across multiple health events may require the merging of several datasets, and databases are incorporating records sourced from different health-care providers \cite{CPRD2019, ukbiobank}. As the volume and dimensionality of EHR data grows, the limitations of existing architectures may become a bottleneck. 

In this study, we implement modifications to the transformer architecture to increase EHR classification performance and make it more suitable to the arising complexity of EHR datasets. Our model, the Bi-Axial Transformer (BAT), learns embeddings that can be utilized for time and sensor attention and preserves signal patterns, enabling it to model complex, multi-axis relationships and the utilize informative missingness, which we demonstrate with visualized attention weights. We quantify how the defining characteristics of EHR datasets, including sparsity, patient demographics, and observed signals, affect performance and give BAT an edge over other transformers. Finally, we look at how the sensor embeddings learned by BAT can be utilized to integrate multiple datasets. Notably, several of the baseline models are implemented with deprecated packages that prohibit a straightforward benchmarking. We re-implemented these models with PyTorch, increased their computational efficiency with a variety of optimizations, and provide them in an open-sourced repository together with BAT at https://github.com/gsn245/Bi-Axial\_Attention\_Transformer. A preliminary version of this work has been reported \cite{devriesparallel}.  

\section{Related Work}\label{sec2}

A variety of deep learning architectures have been utilized for EHR classification, including Convolutional Neural Networks (CNN), Autoencoders, and Graph Neural Networks (GNN) \cite{cheng2016, jun2019stochastic, wang2020feature, zhang2021graph}.
Among the architectures used, recurrent neural networks (RNNs) are some of the most popular models, including variants utilizing Long Short-Term Memory (LSTM) and Gated Recurrent Units (GRU), which frequently achieve top performance \cite{neil2016plstm, saha2023, che2018grud, AYALASOLARES2020, Nasarudin2024}. These models are designed to handle sequential data, with an inductive bias towards recent data points and an ability to handle variable-length inputs. This makes them a natural fit for EHR data, which contains time series of varying lengths due to irregular sampling, and recent observations that may be closer to death or disease onset.

But the nature of digital patient data is evolving, and emerging research is exploring the integration of multiple modalities, such as images and omics, and the harmonization of datasets \cite{NEURIPS2023-0c007ebe, Huang2020, tong2023, biomedicines12071496, marsolo2024assessing}. RNNs often handle new modalities, such as patient metadata, either by projecting it into vector space as the initial hidden state, or by concatenating it to each hidden state before classification. This approach may be unsustainable for longer time series and large modalities, as the architecture is known to have difficulties handling longer input sequences \cite{bengio1994}. These observations highlight the need for new models which can be utilized for the future of EHR classification.    

In recent years, EHR publications have increasingly utilized transformers \cite{li2020behrt, xu2023, Tipirneni2022strats}. Since the introduction of their architecture by \cite{vaswani2017attention}, transformers have been used to explore multidimensional data from novel perspectives, resulting in strong performance in applications such as computer vision and natural language processing (NLP) \cite{vaswani2017attention, dosovitskiy2021visiontransformer}. In comparison to RNNs, transformers show great promise for the future of EHR research, as they perform better on longer sequences, can be parallelized, and can learn richer contextual representations with multi-headed attention \cite{rasmy2020medbert, huang2019clinicalbert, li2020behrt}. There are 2 popular categories of data representation that have been used by transformers for EHR data. The first type represents data points as independent observations in a dense format, often as tuples including the sensor type, time, and measured value. The second type embeds the data as a series of either time points or sensors, condensing multiple observations into a shared embedding that retains signal patterns.

The first strategy is used by models such as SeFT and STraTs \cite{horn2019seft, Tipirneni2022strats}, which are very efficient for highly sparse EHR datasets. However, their dense input representations render them unable to learn from potentially meaningful signal patterns known as informative missingness. 

Informative missingness occurs when the absence of data conveys meaningful information, often reflecting underlying clinical decisions or patient conditions in EHRs. For instance, certain laboratory tests may be performed only when a clinician suspects specific ailments, as highlighted in Beaulieu-Jones \textit{et al.} \cite{jones2018}'s study on structured EHR data. This implies that missing data are frequently not random but instead influenced by factors such as patient demographics, health status, or clinical judgment \cite{jones2018}. Similarly, Che \textit{et al.} \cite{che2018grud} show that the patterns exhibited by the presence and absence of data points strongly correlate with the target label. Because of the clear presence of informative missingness in EHRs, the failure to include sparsity within data representations removes an important source of information. 

 A basic Transformer \cite{vaswani2017attention} is an example of the second type. It can receive a time series input as a collection of temporal embeddings, each of which is learned from all sensor observations, both missing and present, at that time point \cite{thuml2024}. Inversely, the recent iTransformer \cite{liu2023itransformer} was designed for time series forecasting with attention over sensor representations. This resulted in SOTA forecasting for densely observed time series data that improves with increasing lookback windows \cite{zeng2023transformers, das2023long, zhang2022crossformer, ekambaram2023tsmixer}. 

Models in this category preserve informative missingness, but are limited to learning either temporal or intrasensor patterns, which excludes the possibility of discovering more complex sensor-time relationships from the data. This is especially limiting if the embeddings are generated from sparse data, which is predominantly true for EHRs. 

Overall, we contend that both categories of data representation have distinct strengths and weaknesses, highlighting an opportunity for the development of Transformers that can capture complex relationships across both temporal and sensor dimensions, whilst retaining the informative missingness prevalent in EHRs.
\section{Model Development}

\subsection{Problem formulation}
The dataset $\mathcal{D}$ can be defined as a set of tuples, 

\begin{align}
\label{eq:dataset}
\mathcal{D}:= \{(\mX_1, \vp_1, \vh_1, y_1),...,(\mX_N, \vp_N, \vh_N, y_N)\}.
\end{align}

Each sample $i\in N$ has a corresponding multivariate time series of observations $\mX_i$ of size $T_i\times D$, where $x_{i,d}^{t}$ is the observation for sensor $d$ at time $t$. $x_{i,d}$ can be categorical, ordinal, continuous or not observed, depending on $d$. $\vh_i\in\R_{\geq 0}^{T_i}$  is a vector recording the observation times, which vary in length, are irregularly spaced, and are not the same for all $i\in N$. Demographics or observations taken only once at $t=0$ are represented as $\vp_i$, and class labels are indicated as $y\in \{1, ..., C\}$. It is possible that $i$ can have multiple nonexclusive labels, but to simplify the problem we will only handle the case where $|\{y_i\}|=1, \forall i\in\{1, ..., N\}$. 

For EHR data, observations are taken from laboratory tests and vital sign measurements, and $\vp_i$ includes demographic information such as age, sex, and BMI. The objective is to predict class label $y_i$, often representing mortality status, disease diagnosis, or a medical complication, given $(\mX_i, \vh_i, \vp_i)$. This objective is made more challenging by $\mS_i$, which is sparsely sampled for each $i$ across both $t$ and $d$. To represent observed and missing values, a binary indicator matrix $\mM\in\R^{T_i\times D}$ can be generated from $\mX$, with 0 and 1 representing missing and observed values, respectively.

\begin{figure*}[h]
  \centering
    \includegraphics[width=160mm]{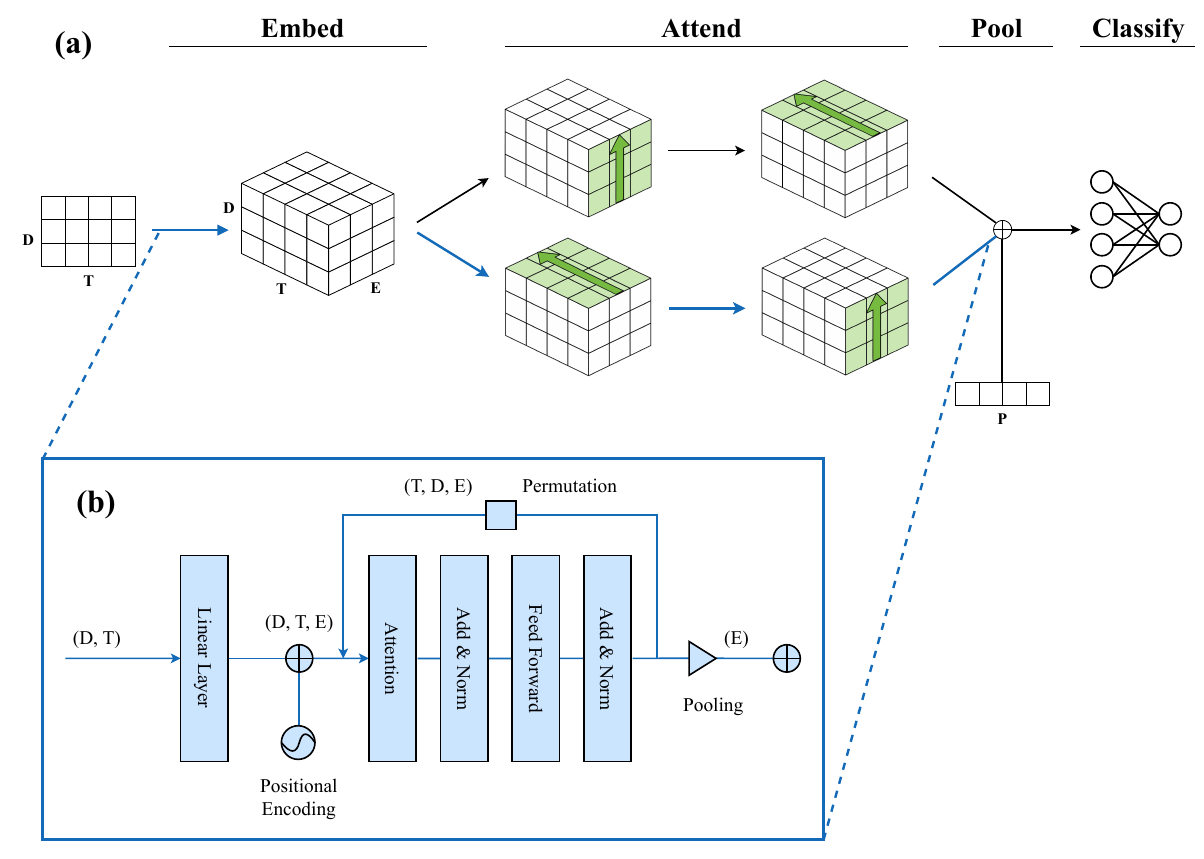}
  \caption{Model architecture for the Bi-Axial Transformer, shown as a) an overview of the architecture and data representation, and b) illustration of an attention track, corresponding with the blue arrows. D represents the sensor identity dimension, T the time dimension, and E the embedding dimension of the time series. P represents demographic data.}
  \label{fig:model_architecture_v1}
\end{figure*}

\subsection{Bi-Axial Transformer Architecture}

BAT is a classification model that has been modified to address two challenges; it preserves the benefits of both data representation strategies introduced in Section \ref{sec2}, and addresses the increasing complexities of EHR data modalities. Specifically, BAT uses a parallelized set of axial attention modules to preserve the benefits of modal and temporal representations demonstrated by other transformer models. It also learns unique sensor embeddings, which may be used across datasets for harmonization, extrapolation, and performance optimization.

The model (Fig. \ref{fig:model_architecture_v1}) can be partitioned into four consecutive stages: 1) embedding the input data, 2) applying attention to the resulting representations, 3) pooling the weighted embeddings, performing classification. We will examine each stage in detail.

\subsubsection{Embedding}
\begin{figure}[!t]
  \centering
  \includegraphics[width=0.6\linewidth]{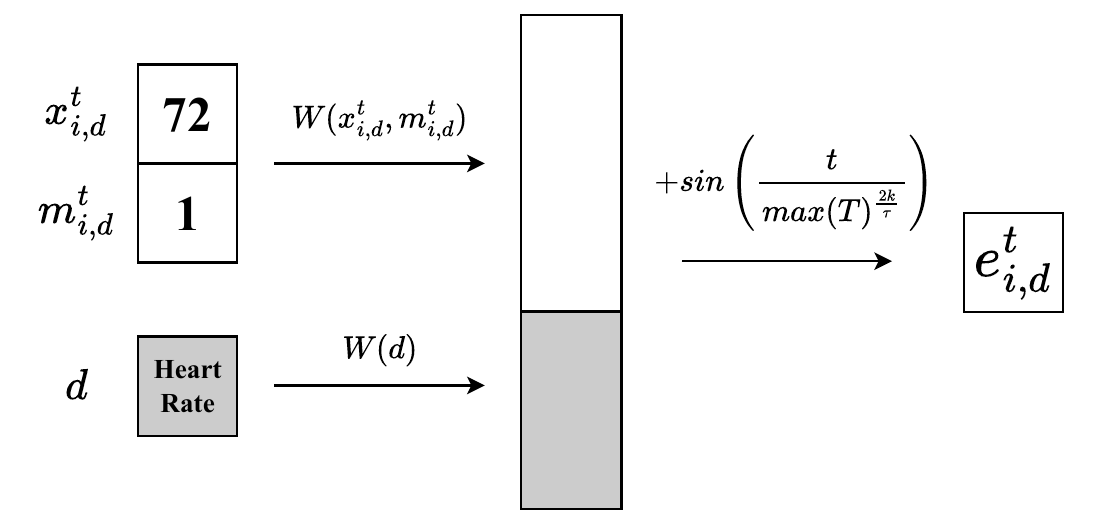}
    \label{fig:subfig1}
  \caption{Example embedding $e$ for an observed heart rate measurement, which is constructed from measured value $x$, binary value $m$ indicating observation occurrence, sensor identity $d$, and a sinusoidal temporal encoding generated from $t$. }
  \label{fig:embedding}
\end{figure}

To apply axial attention to EHR data, initial embeddings must consist of representations that will be meaningful for attention across both axis of $\mX$; that is, between time points and sensors. This is accomplished by generating an embedding for each observation, $e_{i,d}^{t}$, that is constructed from the sensor reading, the sensor identity, and the time of observation as shown in Fig. \ref{fig:embedding}. 

Initially, a linear layer is used to embed sensor readings $\vx_i$, which have been concatenated with $\vm_i \in\mM$ to indicate missingness. All sensor identities $D$ are embedded with a separate linear layer, which will be consistent for all time series and is joined to the sensor reading embedding. Many other models represent $d$ implicitly via a consistent position within $\mX$, but learning unique embeddings for each $d$ enables them to be used for transfer learning on other datasets \cite{che2018grud, shukla2018interpolationprediction}.    

With the observed readings and sensor identities represented, the only component missing from BAT's embeddings are the times. Traditionally, transformers applied to time series adopt positional encodings to incorporate time into embeddings. This makes intuitive sense, considering token order and time are both sequential. Vaswani \textit{et al.} \cite{vaswani2017attention} defined their positional encodings, $PE$, with sine and cosine transformations:

\begin{align}
\label{eq:reg_pe}
PE_k(\text{pos}) = 
\begin{cases} 
\sin\left(\frac{\text{pos}}{10000^{\frac{k}{\tau}}}\right), & \text{if } k \equiv 0 \pmod{2} \\
\cos\left(\frac{\text{pos}}{10000^{\frac{k-1}{\tau}}}\right), & \text{if } k \equiv 1 \pmod{2}
\end{cases}
\end{align}

Here, $pos$ is the position within the sequence, and $10000$ is used as a scaling factor for sine and cosine frequency. $\tau$ represents the encoding dimensions, with $k$ corresponding to even dimensions which utilize the sine function, and odd dimensions $k + 1$ which use cosine. These functions are continuous, allowing the model to generalize to positions not seen during training, while still creating unique encodings for any $pos$. This approach enables increasing input lengths, and makes the functions a natural fit for irregular time series, whose continuous and irregular nature frequently contains unique time points. 

For BAT's time representations, we have adopted a positional encoding used by Horn \textit{et al.} \cite{horn2019seft} to calculate temporal representations:

\begin{align}
\label{eq:seft_pe}
PE_k(\text{t}) = 
\begin{cases} 
\sin\left(\frac{\text{t}}{T^{\frac{k}{\tau}}}\right), & \text{if } k \equiv 0 \pmod{2} \\
\cos\left(\frac{\text{t}}{T^{\frac{k-1}{\tau}}}\right), & \text{if } k \equiv 1 \pmod{2}
\end{cases}
\end{align}

Equation (\ref{eq:seft_pe}) differs from (\ref{eq:reg_pe}), in that the absolute time of observation $t$ replaces the relative sequence position $pos$, and the scaling factor is substituted with the hyperparameter $T$, representing the maximum possible time point. We calculate positional encodings with (\ref{eq:seft_pe}), then add them to the concatentated embeddings. The time series has now been transformed to accommodate a new embedding dimension, $\mX_i\in\R^{T_i\times D\times E}$. 

\subsubsection{Bi-Axial Attention}

BAT's architecture is primarily motivated by 3 assumptions:

\begin{enumerate}
    \item Attention mechanisms are powerful tools for modeling relationships in data.
    \item To fully leverage EHR data, it is essential to capture interdependencies both temporally and across different sensors.
    \item Missing data hold implicit information that can increase predictive power when modeled.
\end{enumerate}

Briefly, self-attention quantifies relationships between the time point representations, generating weights that represent the importance of each to the model's final prediction. An attention layer transforms the input into queries $Q$, keys $K$, and values $V$ with learned weight matrices implemented as linear projections. It then applies an attention mechanism, such as the scaled dot-product:

\begin{align}
\label{eq:scaled dot product}
\mathrm{Attention}(Q,K,V) = \mathrm{softmax}(QK^T/\sqrt{z_k})V
\end{align}

where $z_k$ corresponds to the dimension of each key. The full model will first generate an embedding for each $t$ from all corresponding dynamic features $(x_{i,1}^{t}, ..., x_{i,D}^{t})$. The embeddings are then weighted with self-attention, reduced to a single representation via a pooling layer, and fed to a classification head to predict $y_i$. 

In applying Transformers for time series data we consider the following representation:

\begin{align}
\label{eq:transformer_repr}
\mathbf{E} &= \{ \mathbf{e}_t \in \mathbb{R}^d \mid t = 1, 2, \dots, T \} \\
\mathbf{E}_{inv} &= \{ \mathbf{e}_d \in \mathbb{R}^t \mid d = 1, 2, \dots, D \}
\end{align}

For regular transformers each $e_t$ is an embedding derived from all sensor readings at timepoint $t$, and for inverted transformers each $e_d$ is an embedding derived from the sequence of readings for a specific sensor $d$ across all timestamps $T$.

Both of these types of architectures satisfy assumption \#3 as they retain missingness, but violate assumption \#2 by limiting what we can learn to exclusively temporal or sensor relationships.

Horn \textit{et al.} \cite{horn2019seft} introduce a dense tuple representation of EHR data:

\begin{equation}
\label{eq:seft_reg}
\begin{split}
\mathbf{E}_i = \{\, (t, x_{i,d}^t, d) \mid\ & t \in \{1,\dots,T\},\, d \in \{1,\dots,D\}, \\
                                      & m_{i,d}^t = 1 \,\}
\end{split}
\end{equation}

This representation satisfies assumption \#2 and allows the learning of complex relationships, but it violates assumption \#3 by being a dense representation and only considering sensor and timepoint combinations that have readings. We could incorporate missingness directly in the tuple representation thus giving us:

\begin{equation}
\label{eq:seft_missing}
\begin{split}
\mathbf{E}_i = \{\, (t, x_{i,d}^t, m_{i,d}^t, d) \mid\ & t \in \{1,\dots,T\}, \\
                                              & d \in \{1,\dots,D\} \,\}.
\end{split}
\end{equation}

This approach has the drawback of inflating our sequence size dramatically. Specifically, for our matrix of embeddings $\mathbf{E}$, the number of rows (embeddings required) is:

\begin{align}
\label{eq:number of rows}
|\mathbf{E}|_1 = T \cdot D \cdot (1 - p)
\end{align}

where $p$ represents the overall sparsity of the dataset. If we consider all combinations, regardless of whether they are missing or not (effectively setting $p$ to $0$), it is clear that our sequence length increases dramatically. As attention scales quadratically in the input size, this is an undesirable property. For example, the maximal sequence length for EHR dataset MIMIC-III \cite{johnson2016mimic} would include 2,881 time points and 16 sensors, requiring 46,096 embeddings (Table \ref{dataset_statistics}).

To address these shortcomings and satisfy the assumptions, we employ Axial Attention, a method which is used to address both the vertical and horizontal relationships present in images using embeddings generated from the image channels \cite{DBLP:journals/corr/abs-1912-12180}. These embeddings contains the pixel representation, which alternately attends to all other pixels in its row and column. Analogously, this corresponds to sensors attending to all other readings within its own sensor across all times, and readings across all sensors within the same time. This allows complex relationship learning without requiring a flattened representation, saving a $O(T^{(D-1)/D})$ factor of resources over regular self-attention.

To adapt axial attention for BAT, the data are cloned and passed to two axial attention ``tracks", each of which attend to the time and sensor axes, but in opposite orders. Attention is performed with a bi-directional transformer encoder, which consists of a self-attention head(s), normalization layers and a feed-forward network. For simplicity in the implementation, we do this by transposing either the time or sensor dimension onto the batch dimension, and performing regular self-attention.

This method satisfies hypothesis \#1, as attention can be employed in a transformer model, and also hypothesis \#3 in allowing us to retain missingness in our representation without dramatically increased computational complexity. Hypothesis \#2 is satisfied by sharing encoder parameters across (axis) attention passes within the same track, allowing the model to learn relationships across both sensors and times.

\subsubsection{Pooling, Demographics, and Classification}

Following both rounds of attention, the weighted embeddings from each track are pooled with a mean or max function to reduce dimensionality. The resulting embeddings from both tracks are concatenated and passed through a linear layer and ReLU function for merging. 

In parallel to the attention tracks, BAT also uses the demographic information, learning a separate embedding with a linear layer that is also joined to the learned embeddings and integrated with an additional linear layer and non-linear activation function. The final representation is fed to a two-layer binary classification head. 

BAT is trained on the cross-entropy between binary predictions and patient outcomes:

\begin{align}
\label{eq:scaled dot product}
L = -\frac{1}{N}\sum_{i=1}^Ny_i \cdot \log(p(y_i))+(1-y_i)\cdot \log(1-p(y_i))
\end{align}

With $p(y_i)$ representing the predictions. Model hyperparameters were selected for highest AUROC, following a random sweep with 20 replications for each data set. 

\section{Experiments}

\subsection{Datasets}

Three publicly-available EHR datasets were used to assess BAT on mortality and sepsis classification tasks. 

\textbf{The PhysioNet Challenge 2012 Dataset} (P12) and \textbf{The Medical Information Mart for Intensive Care} (MIMIC-III) are two datasets containing records for ICU stays of approximately 48 hours, with patient mortality labels \cite{goldberger2000physiobank, johnson2016mimic}. All data were preprocessed with a method created by Horn \textit{et al.} \cite{horn2019seft}, which removed outlier samples, one-hot encoded demographic information, and standardized the observations. 

\textbf{The Physionet Challenge 2019 Dataset} (P19) was created for early detection of sepsis \cite{physionet2019}. Each time point is associated with a binary label, indicating whether sepsis occurs at that time or not. We simplify the task by predicting whether or not a patient experienced sepsis during their hospital stay, which leaves us with a single label per patient. Preprocessing was similar to the aforementioned datasets.

\begin{table}
\caption{Statistics for Binary Classification Datasets.}
\label{dataset_statistics}
\centering
\begin{tabular}{p{55pt}p{40pt}p{50pt}p{40pt}p{40pt}p{40pt}}
\toprule
Dataset   & Samples & Max Time \par Points & Sensors & Positive \par Class (\%) & Sparsity \par(\%) \\  \midrule
P12 & 11988 & 215 & 37 & 14.23 & 84.32   \\ 
P19 & 40333 & 336 & 34 & 7.27 & 80.13         \\ 
MIMIC-III & 21107 & 2881 & 16 & 13.22 & 65.50 \\ 
\bottomrule
\end{tabular}%
\end{table}

P12, MIMIC-III, and P19 are all sparse and highly imbalanced, with an underrepresented positive class (Table \ref{dataset_statistics}). To prevent models from only predicting the negative class, each epoch consisted of all positive samples in the training set resampled 3 times, and an equal number of randomly selected negative samples. The datasets were also randomly split into train, test, and validation groups with a ratio of 8:1:1. We repeat this process to generate 5 different datasets, which all models were trained and tested on each for a more robust and comprehensive evaluation. Resulting metrics were averaged across the 5 experiments.

Due to the high sparsity of EHR datasets, a non-EHR dataset was also used to test how varying levels of missingness can impact classification performance. For these experiments, we used the multiclass \textbf{Human Activity Recognition} (HAR) \textbf{Dataset} measured with smartphones \cite{garciagonzalez2020-har}. Sourced from the UC Irvine Machine Learning Repository, each of the 10,299 samples contain readings from a sensor decomposed into 3 acceleration and gravitational measurements, taken at 128 regular time points \cite{human-activity-recognition-using-smartphones-240}. $C = 6$ for this dataset, with each label corresponding to a different physical activity: walking, walking upstairs, walking downstairs, sitting, standing, and laying. 

\subsection{Models} 
Several top-performing models were selected as baselines for our classification experiments. This includes \textbf{GRU-D}, a recurrent neural network which adds a pair of decay mechanisms to a Gated Recurrent Unit \cite{che2018grud}. This helps to control memory retention and the impact of mean sensor values on hidden states, based on the time between consecutive sensor readings.

\textbf{Interpolation-Prediction Networks} (IP-Nets) are comprised of an interpolation network, whose output is given to a prediction network \cite{shukla2018interpolationprediction}. The interpolation network uses kernels to generate interpolants for all possible observations in the time series, $D \times T$. Similarly to GRU-D, IP-Nets also utilize GRUs to make predictions, though the authors clarify that this can be replaced with alternative networks such as LSTMs. 

In order to handle missingness in the input data, both GRU-D and IP-Nets use binary indicator masks to represent data missingness, a strategy which is also used by BAT. 

Unlike the RNNs, \textbf{Set Functions for Time Series} (SeFT) does not include missing observations in their input data. Instead, inputs are represented as time point, sensor, and sensor reading tuples, which are used to generate combined embeddings. This representation is more time and memory-efficient than learning from the entire $D \times T$. The model uses set functions and attention mechanisms on these tuples to aggregate and weigh observations, which are then used to generate predictions.  

Two types of transformer models, the original \textbf{Transformer} by \cite{vaswani2017attention} and \textbf{iTransformer} by \cite{liu2023itransformer} were also implemented for baseline tests. They were modified to contain the same masking components and embedding methods as BAT, to better examine the impact of the axial attention ``tracks" on model performance. 

Hyperparameters for all baselines were selected using a random sweep for each dataset. BAT and all transformer models were trained with a cross-entropy loss function and an AdamW optimizer. Further details regarding the hyperparameters and values swept is provided as supplementary information.

\begin{table*}[h!]
  \centering
  \caption{Mortality Classification from Physionet 2012, MIMIC-III, and Physionet 2019. Best performance is bolded, and second best is italicized.}
  \label{mortality-table}
  \setlength{\tabcolsep}{4pt}
  \begin{tabular}{lcc|cc|cc}
    \toprule
    & \multicolumn{2}{c|}{P12} & \multicolumn{2}{c|}{MIMIC-III} & \multicolumn{2}{c}{P19} \\
    \cmidrule(lr){2-3} \cmidrule(lr){4-5} \cmidrule(lr){6-7}
    Model       & AUPRC $\uparrow$    & AUROC $\uparrow$    & AUPRC $\uparrow$    & AUROC $\uparrow$    & AUPRC $\uparrow$    & AUROC $\uparrow$  \\
    \midrule
    SeFT        & 54.83 $\pm$ 1.92    & 86.05 $\pm$ 0.94    & \textit{52.55 $\pm$ 3.92}  & 84.52 $\pm$ 0.90    & 74.37 $\pm$ 1.70    & \textit{94.83 $\pm$ 0.54} \\
    GRU-D       & \textbf{56.16 $\pm$ 1.94}  & \textbf{87.36 $\pm$ 0.51}  & \textbf{53.07 $\pm$ 2.37}  & \textbf{85.78 $\pm$ 0.79}  & 70.63 $\pm$ 2.68    & 94.36 $\pm$ 0.37 \\
    IP-Nets     & 55.20 $\pm$ 2.22    & 86.30 $\pm$ 0.92    & 51.91 $\pm$ 2.38    & \textit{85.22 $\pm$ 0.83}  & 71.45 $\pm$ 3.34    & 94.69 $\pm$ 0.51 \\
    Transformer & 53.18 $\pm$ 2.08    & 86.42 $\pm$ 0.78    & 52.09 $\pm$ 4.31    & 85.02 $\pm$ 0.37    & \textit{74.73 $\pm$ 1.58}  & 94.79 $\pm$ 0.52 \\
    iTransformer& 49.51 $\pm$ 3.78    & 84.17 $\pm$ 0.94    & 42.35 $\pm$ 2.75    & 80.14 $\pm$ 1.02    & 65.68 $\pm$ 3.41    & 92.32 $\pm$ 0.77 \\
    BAT         & \textit{55.24 $\pm$ 1.45}  & \textit{86.87 $\pm$ 0.86}  & 52.09 $\pm$ 2.38    & 84.50 $\pm$ 0.74    & \textbf{77.05 $\pm$ 1.92}  & \textbf{95.41 $\pm$ 0.58} \\
    \bottomrule
  \end{tabular}
\end{table*}

\subsection{Mortality Prediction}  

\subsubsection{Baseline Comparison} 

BAT and all baseline models were trained and tested for binary mortality classification on P12 and MIMIC-III, and for sepsis on P19. Performance for all tasks were measured with area under the precision-recall curve (AUPRC) and area under the receiver operating characteristic curve (AUROC) (Table \ref{mortality-table}). 
 
It should be noted that modifications were made to IP-Nets and GRU-D in order to boost classification performance for P19. The occurrence of sepsis in the dataset may happen anywhere within ${T}$, and a patient with a positive sepsis label may have had the condition addressed during the hospital stay, and no longer exhibit symptoms at the end of the time series. Because RNNs handle time series sequentially, classifying patients from the final hidden state may result in missed diagnoses. To prevent this error, the hidden states of all time points for P19 were pooled prior to classification. All other datasets were handled in alignment with the original model implementations.

As reported in Table \ref{mortality-table}, BAT performs better on average than the other transformer models, including an average 8.95-point increase over iTransformer for AUPRC, and an average 1.46-point increase 
 over Transformer for the same metric. Transformer matches BAT for MIMIC-III AUPRC and outperforms for AUROC, but a high AUPRC standard deviation indicates that the model is not robust across splits.

Based on model rankings, GRU-D is the best overall, with top performance on both metrics for P12 and MIMIC-III. However, BAT shows a 2.32-point AUPRC and 0.58-point AUROC improvement over the second-best model for P19, the largest classification improvement on any dataset in Table \ref{mortality-table}. Outside of these results, the relative model rankings vary across datasets, and performance may be subject to hyperparameter initialization and dataset splits, as evidenced by high standard deviations. 

This lack of consistency for model ranking can also be found in related literature. Many papers including baseline comparisons for EHR classification have used some or all of these datasets and models, and results across them frequently differ \cite{shukla2018interpolationprediction, horn2019seft}. For method development, this highlights the importance of benchmarking across several dataset splits for a better informed assessment of model performance, as well as the need for up-to-date repositories for benchmarking, which we have created.

\subsubsection{Visualizing BAT Attentions} 

\begin{figure}[h]
  \centering
    \includegraphics[width=88mm]{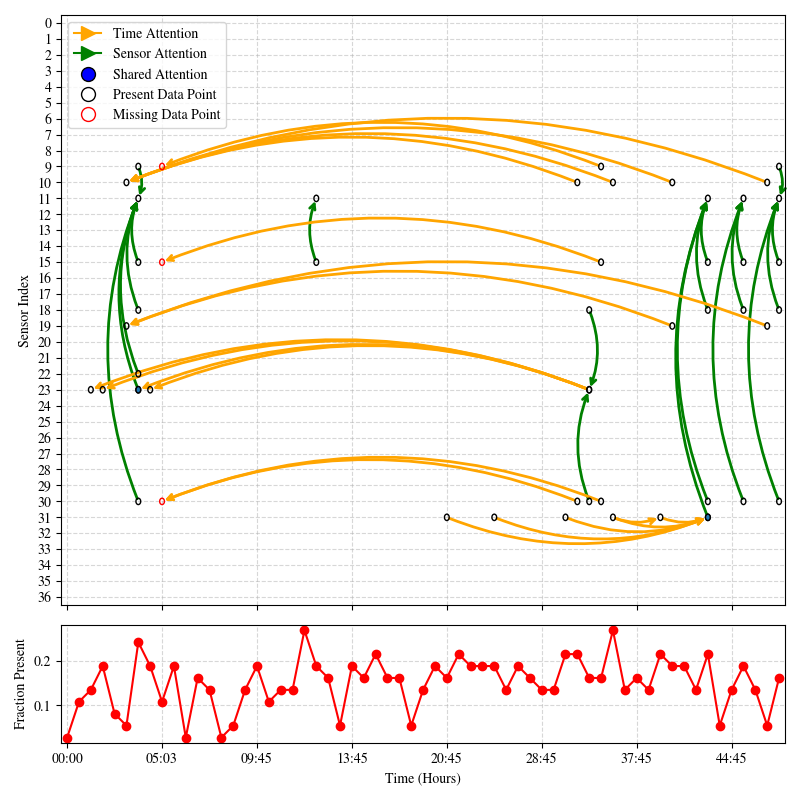}
  \caption{(Top) Readings included in the highest 20 time and sensor attention weights for a randomly selected patient. Sensor attention was performed, followed by time attention. Circles identify the sensor and time of the reading, and arrows indicate the direction of attention from source to the target (i.e., that the target attended strongly to the source). (Bottom) The density of sensor observations at all time points, aligned with the attention weight visualization.}
  \label{fig:attentions_visualized}
\end{figure}

To gain insight into how BAT learns from EHRs, the top 20 time and sensor attention weights from a single track were visualized for a randomly selected P12 patient, who was correctly classified as positive for mortality (Fig. \ref{fig:attentions_visualized}). 

We see that there are embeddings which attend strongly to other sensors at the same time point, which are subsequently attended to strongly by the embeddings from the same sensor at a different time point. These intersectional data points are evidence that BAT passes information between the two axes and that complex relationships spanning both time and sensors are being used by the model to make decisions. Additionally, we see observed data points attending strongly to missing data points across the time axis, suggesting that the absence of readings can be a signal in and of itself. This further suggests that informative missingness influences model decisions.
Further examples and analysis of bi-axial attention in patients can be found in the supplementary section.

\subsection{Transformer Comparisons}

We next performed a series of analyses and experiments to more closely compare BAT to the other transformer models.

\subsubsection{Data Ablations}

To quantitatively analyze the impact that informative missingness and other data comprising $\mathcal{D}$ have on classification, we perform a series of dataset ablations with P12. For these experiments, we maintained the hyperparameters used in experiments from Table \ref{mortality-table}. 

\begin{figure}[htbp]
  \centering
  \begin{subfigure}[t]{0.48\linewidth}
  \centering
    \includegraphics[width=\linewidth]{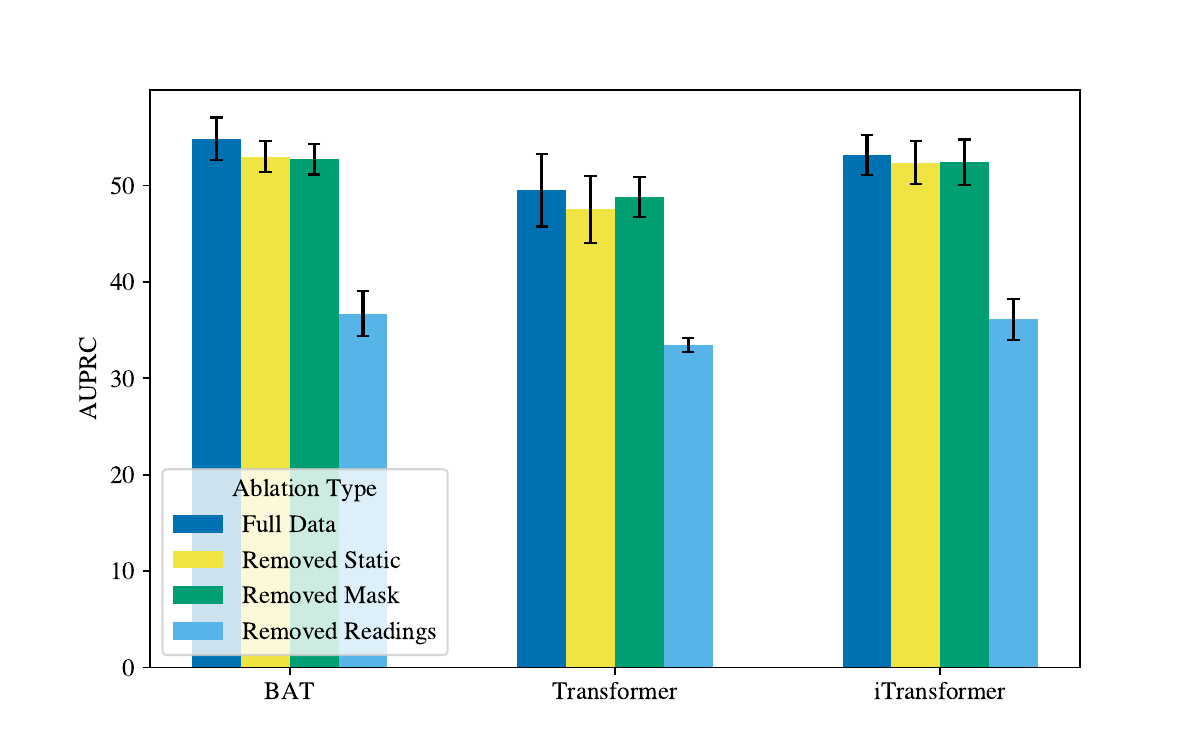}
    \caption{}
    \label{fig:ablation1}
  \end{subfigure}
  \begin{subfigure}[t]{0.48\linewidth}
  \centering
    \includegraphics[width=\linewidth]{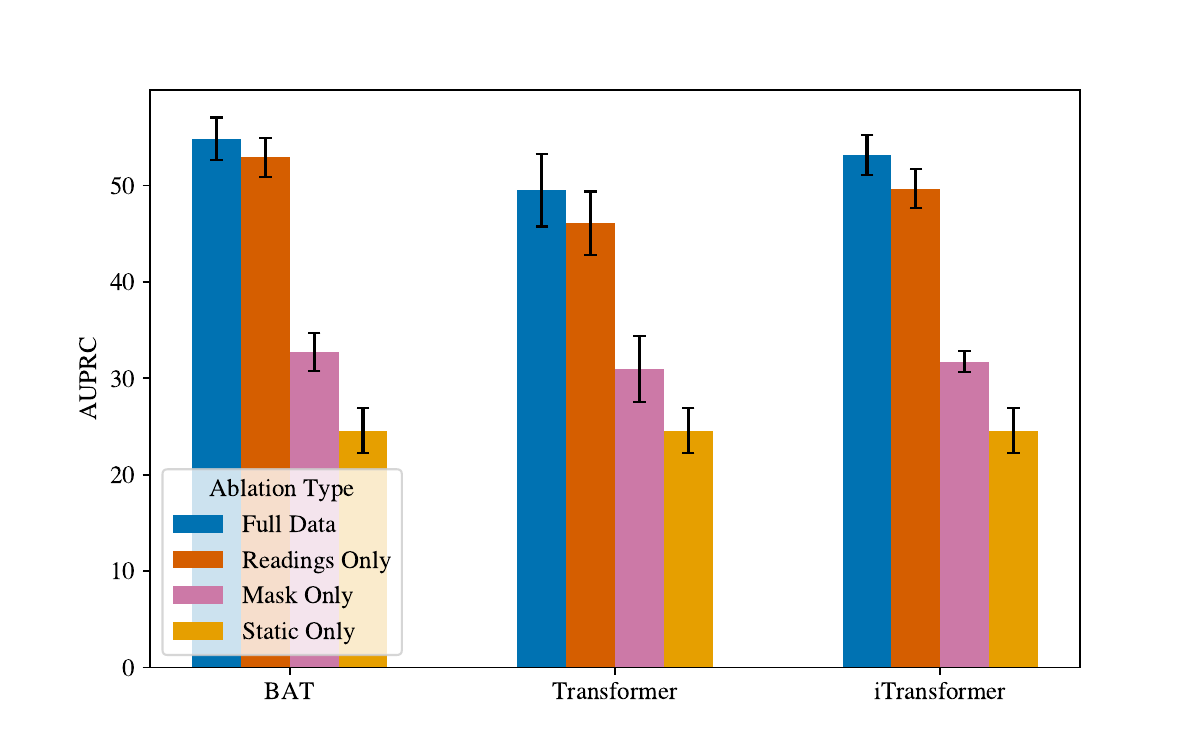}
    \caption{}
    \label{fig:ablation2}
  \end{subfigure}
  \caption{P12 classification with ablations. Includes averages and standard deviations across 5 splits for a) AUPRC after removing data components, and b) AUPRC after removing all but one data component (inverted ablation).}
  \label{fig:ablations_auprc}
\end{figure}

Fig. \ref{fig:ablations_auprc} presents the results as two sets of ablations. Fig. \ref{fig:ablation1} follows the format of a traditional ablation, where one part of the dataset $\mathcal{D}$, such as the indicator mask $M$, is removed, and the model is trained on the remaining data. Fig. \ref{fig:ablation2} are the results an inverted ablation, to test how well the models can classify patients based only on one component of the data. For example, providing exclusively $M$ to the models tests their ability to learn only from informative missingness, as no sensor readings, demographics, or any information besides signal patterns are present.

For all transformer models, the entirety of the dataset is needed to achieve optimal performance, though any ablation that includes the sensor readings achieves close results. Relatedly, it can be seen in Fig. \ref{fig:ablation1} that removing the sensor readings unanimously results in the largest performance reduction. This indicates that models learn the most information from $X$ in order to make a classification. However, models trained on any data subset performs better than random, which would result in an AUPRC equal to the ratio of positive samples (Table \ref{dataset_statistics}). 

Fig. \ref{fig:ablation1} also shows that removing the demographic information and indicator mask affects models differently. Both BAT and Transformer are similarly impacted by each ablation, but iTransformer has a lower performance after removing the static data than after removing the indicator mask. This may be due to a larger dependence on demographic information, or a larger overlap in the information that can be learned from the binary indicator mask and sensor readings.  

The better-than-random performance of the indicator mask-only classification in Fig. \ref{fig:ablation2} provides further evidence for informative missingness. Importantly, BAT has a higher resulting AUPRC on any one subset of the data than the other models.  The static-only test results are the same for all models, as the demographic representations would be learned without attention, thus being produced by effectively the same model. 

Similar patterns can be found for AUROC, as seen in the supplementary information.

\subsubsection{Sparsity}

\begin{figure}[htbp]
  \centering
  \begin{subfigure}[t]{0.48\linewidth}
  \centering
    \includegraphics[width=\linewidth]{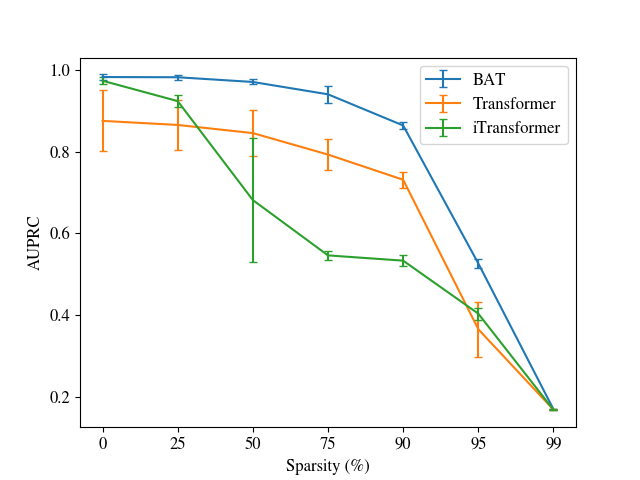}
    \caption{}
    \label{fig:subfig1}
  \end{subfigure}
  \begin{subfigure}[t]{0.48\linewidth}
  \centering
    \includegraphics[width=\linewidth]{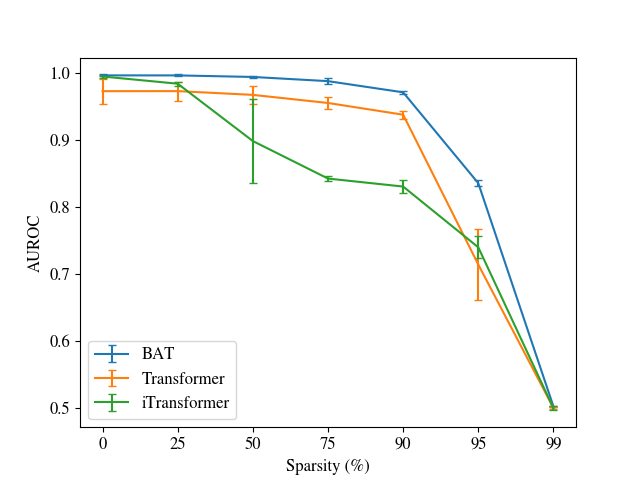}
    \caption{}
    \label{fig:subfig2}
  \end{subfigure}
  \caption{Multiclass a) AUPRC and b) AUROC of transformer models trained and tested on HAR \cite{garciagonzalez2020-har}. Averages and standard deviations are calculated from 5 dataset splits.}
  \label{fig:sparsity_line_graph}
\end{figure}

 Another defining feature of EHRs is sparsity, which varies across datasets but is frequently high. To compare the transformer models' robustness to sparsity, missingness was induced through random selection and removal of observations in the HAR dataset. This process was applied at seven different sparsity levels: 0, 25, 50, 75, 90, 95, and 99 percent. The multiclass macro AUPRC and AUROC (one-vs.-rest) were then calculated and averaged across 5 different data splits at each sparsity level. 

As seen in Fig. \ref{fig:sparsity_line_graph}, BAT has better performance at all levels of missingness than Transformer and iTransformer, suggesting that the bi-axial attention adds model robustness by sharing information across axes. It is also more stable across all data splits, as evidenced by lower standard deviations. However, there are some limitations to sparsity level and model performance, as all models perform equivalently to random guessing when the dataset is 99\% sparse. 

\subsection{Sensor Representations}

BAT's embedding structure, particularly with regards to the learned sensor embeddings, makes it unique to many other EHR classification architectures. None of the baseline models, which were selected for their performance and popularity in other publications, generate unique sensor embeddings which can be extracted or shared across multiple datasets. GRU-D, IP-Nets, and iTransformer learn sensor identities implicitly from a consistent location within the inputs. SeFT's embeddings are learned jointly from the time point, sensor, and reading, making it impossible to disentangle the sensor embedding. Similarly, the Transformer learns a joint embedding for each timepoint over all sensors. 

With explicit sensor embeddings, BAT has the potential to extend beyond the abilities of other classification models and address a variety of tasks for datasets with overlapping sensors. This includes dataset integration, transfer learning, bias correction, and increased interpretability via a common feature space. This may be particularly valuable for datasets which are smaller, highly sparse, or are missing labels.

To investigate this capability, we trained two versions of BAT on a combined dataset consisting of P12 and MIMIC-III. The first version contained separate sensor embeddings for each dataset, whereas the second used shared embeddings for sensor identities present in both datasets. The models were then evaluated for classification on each dataset. As seen in Table \ref{sensor-embeddings-table}, the shared sensor embeddings result in superior performance on all metrics but AUPRC on MIMIC-III. This may be due to the larger number of outliers in MIMIC-III, or the more complex data distribution created by a larger number of patients with longer time series in comparison to P12. However, it can be noted that the variance for both datasets is greatly reduced by the combined sensor embeddings, showing increased model stability when combining identical sensors.

\begin{table}
\caption{Mortality Classification for Physionet 2012 and MIMIC-III. Trained on joint datasets with overlapping sensors represented with separate or shared embeddings. Best performance is bolded.}

\label{sensor-embeddings-table}
\centering
\begin{tabular}
{p{1cm}p{2cm}p{2cm}p{2cm}p{2cm}}
\toprule
 & \multicolumn{2}{c|}{{P12}} & \multicolumn{2}{c}{{MIMIC-III}}  \\ \midrule
\multicolumn{1}{l}{} & \multicolumn{1}{l}{{AUPRC $\uparrow$}} & {AUROC $\uparrow$} & \multicolumn{1}{|l|}{{AUPRC $\uparrow$}} & {AUROC $\uparrow$} \\ \midrule 
\multicolumn{1}{l}{Separate} & \multicolumn{1}{l}{50.71 +/- 3.11} & 84.22 +/- 2.14 & \multicolumn{1}{|l}{\textbf{45.48 +/- 5.54}}       & {80.15 +/- 1.92} \\
\multicolumn{1}{l}{Shared} & \multicolumn{1}{l}{\textbf{52.07 +/- 1.51}} & \textbf{84.98 +/- 1.33} & \multicolumn{1}{|l}{43.87 +/- 2.18} & \textbf{81.95 +/- 1.62} \\ \bottomrule 
\end{tabular}
\end{table}

\section{Discussion}
We present and evaluate the Bi-Axial Transformer, which attends to both the temporal and clinical sensor axes of EHR data to capture complex relationships and account for data sparsity. BAT is best overall for sepsis classification on P19, with a large improvement over the next-best method. However, performance for BAT and all baselines varied across datasets, highlighting the importance of testing a variety of models, splits, and initializations to find the best method for new datasets. Investigation into model attention maps revealed evidence of BAT learning from informative missingness and complex time-sensor relationships, which are critical components of EHR data and motivate our model architecture.
BAT is also an improvement over transformers in classification performance, and has increased robustness to sparsity in HAR. 

Finally, we illustrate BAT's potential for merging multiple datasets by sharing sensor representations between P12 and MIMIC-III, paving the way for future models that integrate heterogeneous clinical datasets.
Our approach to generating sensor identity embeddings is simple but successful, as a linear layer applied to one-hot encodings. It can be seen, though, that there is room for potential improvements and more sophisticated embedding methods. Future work could explore more advanced embedding strategies, such as knowledge graph-based representations or LLM-derived embeddings informed by clinical literature. Notably, our method is modular and generalizable, allowing it to be integrated into any model that learns sensor identity through a consistent position in the input data.

\section{Conclusion}
Historically, strong progress has been made in modeling EHR datasets as irregular, sparse, and multimodal time series, particularly for classification tasks. But EHRs are continuing to grow in size and complexity, presenting a demand for models that can learn richer data relationships and integrate multiple datasets. We laid out 3 hypotheses for how a bi-axial attention model may be well suited to the task, and achieved performance competitive to or exceeding state of the art in mortality and sepsis prediction tasks. Our visualization and analysis of model attention patterns suggests that these hypotheses were well-founded and that the model uses both time-sensor relationships and informative missingness to produce successful classifications. We also demonstrate that BAT's data representation allows us to combine multiple datasets, indicating potential for large-scale pretraining similarly to language models in the NLP. Overall, we believe that BAT is a useful tool for handling the emerging demands of modern EHR datasets, and further exploration of its potential may enable better classification and improve insights for precision medicine. 

\newpage
\bibliographystyle{unsrtnat}
\bibliography{bibliography_backup}  

\newpage
\section{Supplementary Information}

\vspace{0pt}

\subsection{Hyperparameter Sweep}
\label{hyperparameter-sweep-section}

For optimal model performance, we conducted a hyperparameter sweep that sampled dropout, attention dropout, number of heads, number of layers, pooling function, learning rate, embedding size, and batch size for BAT, Transformer, and iTransformer. For SeFT, GRU-D, and IP-Nets, the sweep included additional architecture-specific parameters, which can be seen in  Table \ref{hyperparameter-table}. All hyperparameter value ranges that were selected from during the sweep can be found in the Github repository, https://github.com/gsn245/Bi-Axial\_Attention\_Transformer. To select the best model, hyperparameter values were randomly sampled 20 times, generating 20 potential model candidates. Each candidate was trained and tested on 5 splits of train, validation, and test data in an 8:1:1 ratio, and the model with the highest resulting AUROC was selected. 

\begin{table}[b!]
\caption{Baseline Model Hyperparameters.}
\label{hyperparameter-table}
\scalebox{.92}{
\begin{tabular}{p{2cm}p{4.5cm}p{4.5cm}p{4.5cm}}
\multicolumn{1}{p{1cm}}{}  &\multicolumn{1}{p{3cm}}{\bf P12} &\multicolumn{1}{p{3cm}}{\bf MIMIC-III}    &\multicolumn{1}{p{3cm}}{\bf P19}
\\ \toprule \\
SeFT             &dropout:0.4, attn\_dropout: 0.3, phi\_dropout: 0.3, rho\_dropout: 0.0, heads: 2, lr: 0.01, phi\_layers: 1, phi\_width: 512, psi\_layers: 5, psi\_width: 32, psi\_latent\_width: 128, dot\_prod\_dim: 512, latent\_width: 64,  rho\_layers: 2, rho\_width: 256, batch: 128,  max\_timescale: 1000, positional\_dims: 16 
&dropout:0.4, attn\_dropout: 0.1, phi\_dropout: 0.3, rho\_dropout: 0.0, heads: 3, lr: 0.001, phi\_layers: 3, phi\_width: 256, psi\_layers: 2, psi\_width: 64, psi\_latent\_width: 16, dot\_prod\_dim: 512, latent\_width: 256,  rho\_layers: 3, rho\_width: 128, batch: 32,  max\_timescale: 10, positional\_dims: 8 
&dropout:0.4, attn\_dropout: 0.3, phi\_dropout: 0.0, rho\_dropout: 0.0, heads: 2, lr: 0.001, phi\_layers: 2, phi\_width: 32, psi\_layers: 2, psi\_width: 256, psi\_latent\_width: 32, dot\_prod\_dim: 256, latent\_width: 512,  rho\_layers: 2, rho\_width: 256, batch: 32,  max\_timescale: 100, positional\_dims: 8
\\ \midrule \\
GRU-D             &dropout:0.4, recurrent\_dropout:0.2, n\_units:128, lr:0.0001, batch:32 
&dropout:0.0, recurrent\_dropout:0.4, n\_units:100, lr:0.001, batch:256 
&dropout:0.3, recurrent\_dropout:0.4, n\_units:100, lr:0.001, batch:128\\ \midrule \\
IP-Nets             &dropout:0.0, recurrent\_dropout:0.3, lr:0.001, batch:32, n\_units:32, impute\_stepsize:1, reconstruct\_fraction:0.75 
&dropout:0.0, recurrent\_dropout:0.3, lr:0.001, batch:256, n\_units:100, impute\_stepsize:0.5, reconstruct\_fraction:0.005 
&dropout:0.4, recurrent\_dropout:0.3, lr:0.0001, batch:16, n\_units:100, impute\_stepsize:1, reconstruct\_fraction:0.2 
\\ \midrule \\
Transformer             &dropout:0.2, attn\_dropout:0.1, heads:1, layers:3, pool:max, lr:0.0001, embed\_size:8, batch:16 
&dropout:0.1, attn\_dropout:0.3, heads:1, layers:3, pool:max, lr:0.001, embed\_size:8, batch:32 
&dropout:0.1, attn\_dropout:0.2, heads:1, layers:3, pool:max, lr:0.0001, embed\_size:8, batch:16 \\ \midrule \\
iTransformer             &dropout:0.4, attn\_dropout:0.4, heads:2, layers:3, pool:max, lr:0.001, embed\_size:8, batch:16 
&dropout:0.2, attn\_dropout:0.2, heads:2, layers:3, pool:mean, lr:0.01, embed\_size:8, batch:32 
&dropout:0.1, attn\_dropout:0.1, heads:3, layers:1, pool:mean, lr:0.01, embed\_size:8, batch:16 \\ \midrule \\
BAT             &dropout:0.1, attn\_dropout:0.4, heads:2, layers:1, pool:max, lr:0.0001, embed\_size:128, batch:16 
&dropout:0.4, attn\_dropout:0.2, heads:1, layers:1, pool:mean, lr:0.0001, embed\_size:16, batch:4
&dropout:0.4, attn\_dropout:0.4, heads:1, layers:1, pool:max, lr:0.0001, embed\_size:32, batch:16
\\ \bottomrule \\
\end{tabular}
}
\end{table}

\subsection{Additional Data Ablations}

\begin{figure}[htbp]
  \centering
  \begin{subfigure}[t]{0.48\linewidth}
  \centering
    \includegraphics[width=\linewidth]{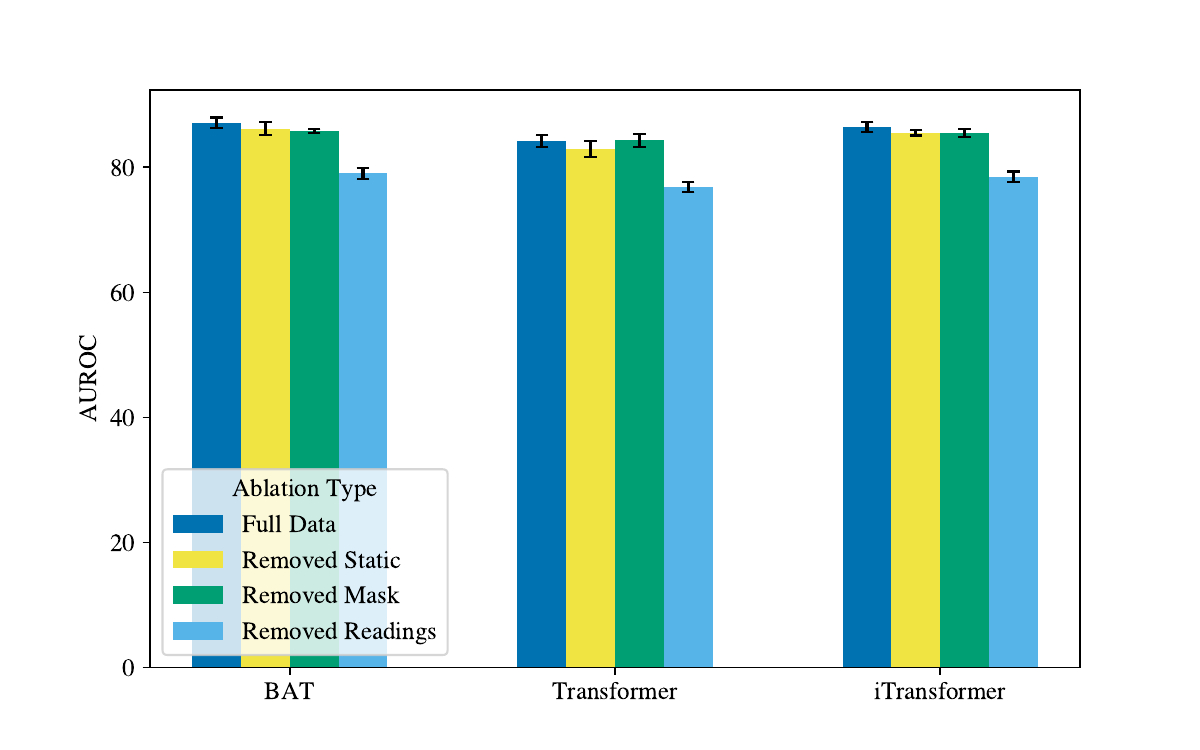}
    \caption{}
    \label{fig:subfig1}
  \end{subfigure}
  \begin{subfigure}[t]{0.48\linewidth}
  \centering
    \includegraphics[width=\linewidth]{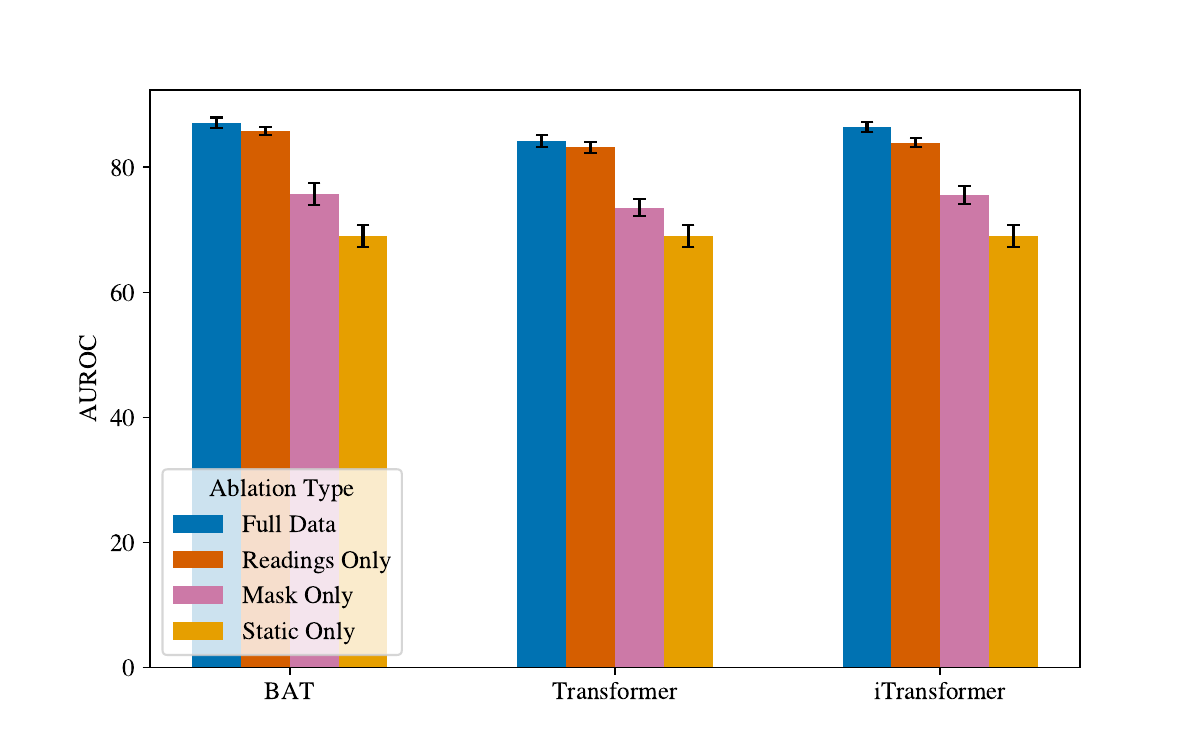}
    \caption{}
    \label{fig:subfig2}
  \end{subfigure}
  \caption{P12 classification with ablations. Includes averages and standard deviations across 5 splits for a) AUROC after removing data components, and b) AUROC after removing all but one data component (inverted ablation).}
\label{fig:auroc_ablations}
\end{figure}

In addition to determining the AUPRC for data ablations on Transformer, iTransformer, and BAT for P12, we also calculated the AUROC for the same experiments (Fig. \ref{fig:auroc_ablations}). The impact of each data component, or its absence, is similar in pattern to AUPRC, with a smaller decrease in performance relative to the full data. For example, the static only ablations in Fig. \ref{fig:subfig2} have 68.96 AUROC, which is approximately 79\% of BAT's AUROC for the full data, 86.87, whereas the AUPRC for the same tests resulted in a larger performance decrease from 55.24 to 24.57.

\end{document}